\documentclass[letterpaper, 10 pt, journal, twoside]{IEEEtran} 
\usepackage{algorithm} %format of the algorithm
\usepackage{algorithmicx}  
\usepackage{algpseudocode}
\usepackage{tabularx}

\IEEEoverridecommandlockouts      % This command is only needed if 
\usepackage{graphics} % for pdf, bitmapped graphics files
\usepackage{epsfig} % for postscript graphics files
\usepackage{mathptmx} % assumes new font selection scheme installed
\usepackage{times} % assumes new font selection scheme installed
\usepackage{amsmath} % assumes amsmath package installed
\usepackage{amssymb}  % assumes amsmath package installed
\usepackage{algpseudocode}
\usepackage{verbatim}
\usepackage{booktabs}
\usepackage{graphicx}
\usepackage{epsfig}
\usepackage{cite}
\usepackage{tensor}
\usepackage{color}
\usepackage{bm}
\usepackage{graphicx} %加入头文件
\usepackage{fancyhdr} % 添加页眉页脚
\graphicspath{{figures/}}
\DeclareGraphicsExtensions{.pdf}
\usepackage{mathtools}
\usepackage{algorithm} %format of the algorithm
\usepackage{threeparttable}
\usepackage{multirow} %multirow for format of table
\usepackage{amsmath}
\usepackage{xcolor}
\usepackage[colorlinks,linkcolor=red,anchorcolor=blue,citecolor=green]{hyperref}
\title{\LARGE \bf
A Robotic Line Scan System with Adaptive ROI for Inspection of Defects over Convex Free-form Specular Surfaces
}

\author{Shengzeng Huo, David Navarro-Alarcon and David Chik%

\thanks{S. Huo and D Navarro-Alarcon are with The Hong Kong Polytechnic University, KLN, Hong Kong. Corresponding author: {\texttt{\small dna@ieee.org}}.}%
\thanks{D. Chik is with the Hong Kong Applied Science and Technology Research Institute, NT, Hong Kong.}%
\thanks{This work is supported in part by ASTRI under grant number PO2019/399, in part by the Research Grants Council (RGC) of Hong Kong under grant number 14203917 and in part by PolyU under grant G-YBYT.}%
}

\begin{document}

\maketitle
\thispagestyle{fancy}
\cfoot{\thepage} % 页眉中
\renewcommand{\headrulewidth}{0pt} %改为0pt即可去掉页眉下面的横线
\renewcommand{\footrulewidth}{0pt} %改为0pt即可去掉页脚上面的横线

\pagestyle{fancy}
\cfoot{\thepage}

%%%%%%%%%%%%%%%%%%%%%%%%%%%%%%%%%%%%%%%%%%%%%%%%%%%%%%%%%%%%%%%%%%%%%%%%%%%%%%%%
\begin{abstract}
In this paper, we present a new robotic system to perform defect inspection tasks over free-form specular surfaces. The autonomous procedure is achieved by a six-DOF manipulator, equipped with a line scan camera and a high-intensity lighting system. Our method first uses the object’s CAD mesh model to implement a K-means unsupervised learning algorithm that segments the object’s surface into areas with similar curvature. Then, the scanning path is computed by using an adaptive algorithm that adjusts the camera’s ROI to observe regions with irregular shapes properly. A novel iterative closest point-based projection registration method that robustly localizes the object in the robot’s coordinate frame system is proposed to deal with the blind spot problem of specular objects captured by depth sensors. Finally, an image processing pipeline automatically detects surface defects in the captured high-resolution images. A detailed experimental study with a vision-guided robotic scanning system is reported to validate the proposed methodology. 
\end{abstract}
% \begin{IEEEkeywords}
% Robotics; Line Scan; Defect Inspection; Motion Control
% \end{IEEEkeywords}
%%%%%%%%%%%%%%%%%%%%%%%%%%%%%%%%%%%%%%%%%%%%%%%%%%%%%%%%%%%%%%%%%%%%%%%%%%%%%%%%
\section{INTRODUCTION}
Surface inspection is important in modern manufacturing because defects severely affect the value and quality of products. The common practice of many factories is still to hire numerous workers to perform the inspection task manually, which is costly and time-consuming, and the performance can be very subjective. Many automatic inspection systems have been developed by using machine vision algorithms and high-resolution cameras to solve these problems. Several typical applications are inspection of railway surfaces \cite{girardeau2015cloud}, automotive exterior body parts \cite{tandiya2018automotive}, and steel surfaces \cite{neogi2014review}.

\begin{figure}
    \centerline{\includegraphics[width=\columnwidth]{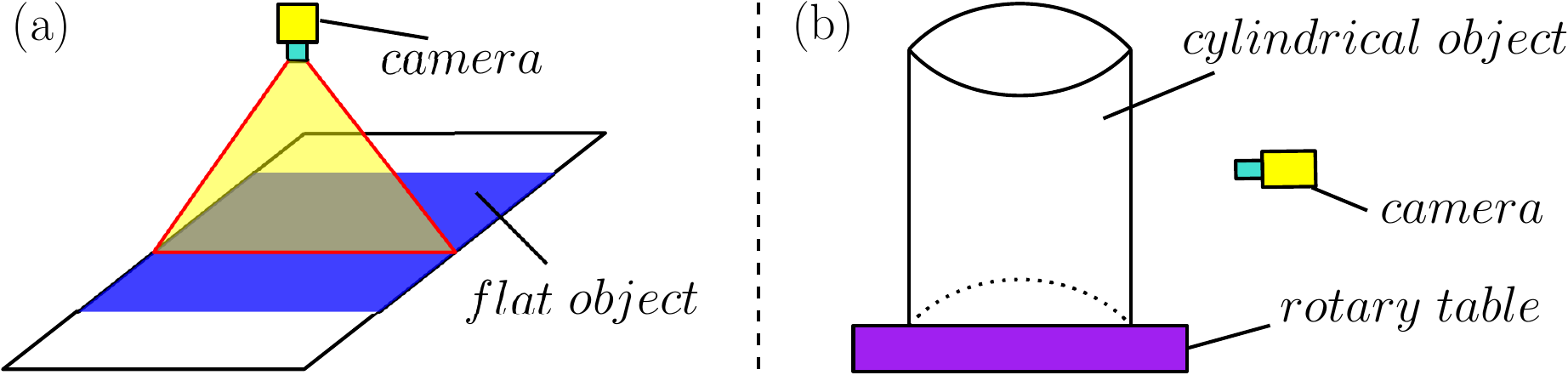}}
    \caption{Two typical inspection setups in production line: (a) Mount the camera on top and capture continuous flat plane, (b) Capture images of different angles with a object on a rotary table.}
    \label{inspection setup}
\end{figure}

A typical setup of an automatic line camera inspection system in a production line is shown in Fig. \ref{inspection setup}(a), where a camera is mounted on top of the workspace to scan objects continuously. Many previous works developed these types of systems. \cite{cho2005development} proposed a fast fabric inspection system focusing on objects with a continuous stream, such as paper and plywood. \cite{fernandez1993vision} used matrix cameras instead of linear ones with 1 $mm^2$ resolution to deal with continuous flat metallic products in the production line. \cite{fan2015automatic} adopted a diffusion-lighting system to avoid the strong reflections from the surface of car's body. Using a rotating table with a fixed scan camera is another common setup, as shown in Fig. \ref{inspection setup}(b). The camera can capture multiple images of parts from different angles with this setting. \cite{RenweiHu2018Design} used an industrial camera and coaxial area lighting to scan smooth products with an electronically controlled rotary table. \cite{zhang2015line} used a high-resolution line camera to scan cultural heritage objects, with a table that rotates 5 degrees after each scan.

The common feature of the two above approaches is the use of a fixed camera with a moving object, which is only suitable for certain standard shaped objects. 
However, free-form parts are very common in industrial production. Furthermore, high-intensity light should be precisely aligned with the camera when scanning specular surfaces. However, the two above approaches cannot deal with these issues. An industrial camera with a machine vision lighting system is equipped with a robot manipulator in our system to overcome the drawbacks, enabling moving around the object to scan all sides. According to \cite{freeman2012machine}, the characterization of free-form surfaces is defined as a well-defined surface normal that is continuous almost everywhere except at vertices, edges, and cusps. Considering the camera settings and real inspection conditions, we concentrate on convex free-form objects in this paper, with detailed understanding provided by \cite{les2003shape}. \par

In this paper, we focus on defect inspection issues over free-form specular surfaces. High reflection ratio and continuous variation of curvature are two main challenges in this task. For the former, the clean CAD mesh model of the object is used to guide our scanning path planning. In addition, a projection boundary registration algorithm for specular surfaces with the help of an RGB-D camera is proposed to localize the object in robot's frame. For the latter, the free-form CAD model is divided into several flat regions to fit the scanning range of the camera. The main original contributions of this paper are as follows:
\begin{itemize}
    \item Provide a complete pipeline to plan the path of the line scanner by K-means region segmentation and an adaptive ROI algorithm. 
    \item Design an iterative closest point(ICP)-based projection boundary registration method for specular objects.
    \item Validate the developed automatic inspection system with experiments using free-form specular objects. 
\end{itemize}
The remainder of this paper is organized as follows. Sec. 2 introduces the architecture of the system. Sec. 3 describes the scanning path planning. Sec. 4 presents the sensing and control part. Sec. 5 shows the experimental setup and results. Finally, Sec. 6 provides the conclusion.\par

\section{ROBOTIC SYSTEM}
For an outstanding defect inspection machine vision system, highlighting defects and sufficient detection resolution are two fundamental requirements. Applying area scan camera into inspection application has several disadvantages, such as blurring \cite{cho2005development}. By constrast,  the high-resolution, high-frequency characteristics of line scan camera make it ideal for scanning surfaces with continuously varying curvature \cite{onodera2013high}. Thus, a line scan camera, Basler raL2048-48gm GigE camera, is used in our system. Generally, lighting is always needed in the field of machine vision to enhance the difference between the defects and the background. Focusing on specular surfaces, dark-field illumination is always adopted \cite{pham2002smart}, in which only the light encountering the defects passes into the camera. Thus, line lighting is used to fit the line scan camera. 

\begin{figure}
    \centerline{\includegraphics[width=\columnwidth]{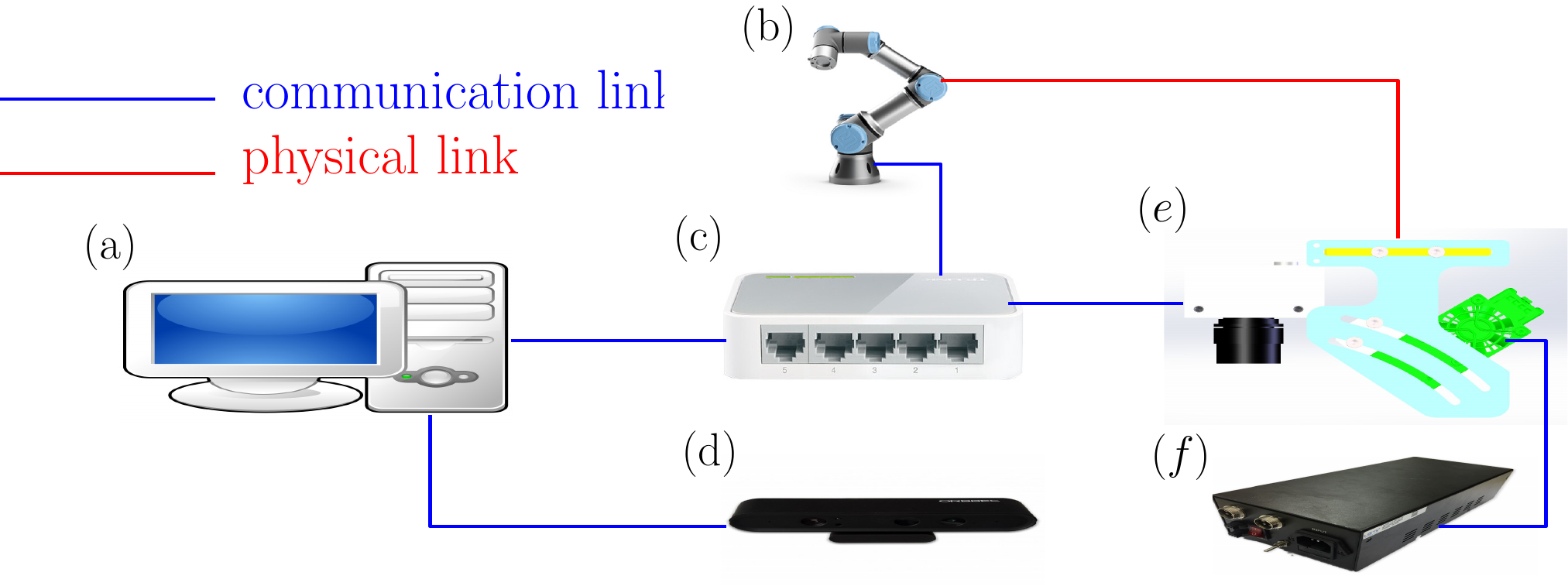}}
    \caption{Architecture of the proposed robotic line scan system. (a) Linux PC. (b) UR3 robot manipulator. (c) Network router. (d) Orbbec Astra S depth camera. (e) Image acquision system, including line scan camera, line light and a holder. (f) Analog control box high power LED strobe. }
    \label{system_architecture}
\end{figure}

\begin{figure}
    \centerline{\includegraphics[width=\columnwidth]{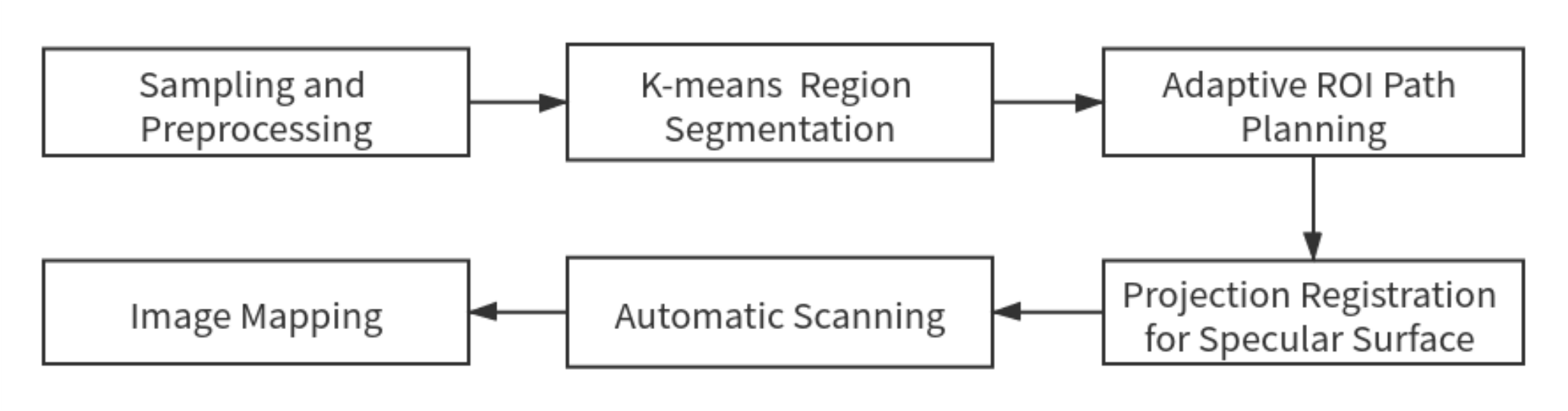}}
    \caption{Complete defects detection pipeline}
    \label{Pipeline}
\end{figure}

Fig. \ref{system_architecture} illustrates the architecture of our proposed robotic prototype. A UR3 manipulator from Universal Robots is used to achieve a six-DOF path. The line lighting and line scan camera are equipped with the end-effector of the UR3 manipulator. Our system also includes an Orbbec Astra RGB-D camera to localize the object in the robot's frame by registration with the CAD mesh model. A network router and ROS \cite{quigley2009ros} are used to establish a local network and communicate different components in the system. The complete system can be divided into two main subsystems: a planning system as well as a sensing and control system. The pipeline of the entire system is shown in Fig. \ref{Pipeline}. 

\section{PLANNING SYSTEM}
\emph{Notation.} Throughout this paper, we use very standard notation. Column vectors are denoted with bold small letters $ \vec{{\mathbf v}}$ and matrices with bold capital letters $\mathbf M$. 

\subsection{Sampling and Preprocessing} 
\begin{figure}
    \centerline{\includegraphics[width=\columnwidth]{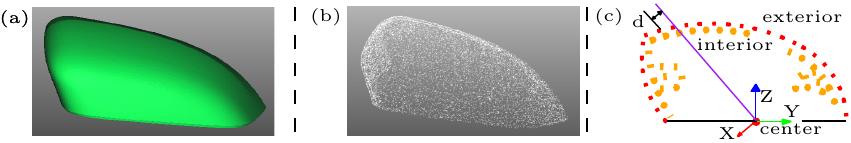}}
    \caption{ Preprocessing for input mesh model: (a) Typical free-form CAD mesh model. (b) Sampled point cloud from mesh model. (c) Principle of filtering irrelevant points.}
    \label{preprocessing}
\end{figure}

\cite{girardeau2015cloud} provided a useful tool to sample points from mesh model by randomly selecting a given number of points on each triangle. The typical mesh model and sampled point cloud $M$ are shown in Fig. \ref{preprocessing}(a) and (b), respectively. Each point in $M$ is a vector $ {\Vec{\mathbf m}}={[x \ y\ z]}^T$. Generally, only the exterior surface of the object is emphasized in defect inspection of quality control. However, the sampled point cloud includes points from all faces of the object. The irrelevant points are not only useless for our scanning, but also will affect the normal estimation in the following steps. As a result, they must be filtered out.\par
Here, irrelevant points are defined as the points closer to the center of the object. The center of point cloud $M$ can be computed by
\begin{equation}
 {\Vec{\mathbf c}}=\sum_{i}^{N_M}{ {\Vec{\mathbf m}_i}}/N_M
\label{center_point}
\end{equation}
where $ \vec{\mathbf m}_i$ is i-th point of $M$, and $N_M$ is the size of $M$. Our solution is to eject multiple straight lines from $\Vec{\mathbf c}$ to the surface in different directions and only save the farthest point among intersection points in each line, as shown in Fig. \ref{preprocessing}(c). A sphere coordinate is used to distribute the lines evenly.
\begin{equation}
\label{line ejection}
 \Vec{\mathbf l}_{\theta,\phi} ={[r\sin{\theta}\cos{\phi}\ r\sin{\theta}\sin{\phi}\ r\cos{\theta}]}^T
\end{equation}
where $r,\theta,\phi$ represent radius, polar angle, and azimuthal angle, respectively. 
The points whose distance to the line is smaller than a threshold are considered on the line. The distance between points and lines can be computed by 
\begin{equation}
d=\frac{ {\Vec{\mathbf q}_i}\cdot \Vec{\mathbf l}_{\theta,\phi} } {\left| {\Vec{\mathbf q}_i} \right|\cdot\left| {\Vec{\mathbf l}_{\theta,\phi}}\right|}
.\label{point_to_line}
\end{equation}
% \begin{equation}
% d={\Vec{\mathbf q}_i}\cdot \Vec{\mathbf l}_{\theta,\phi}  /(\left| \Vec{\mathbf q}_i \right| \cdot\left| \Vec{\mathbf l}_{\theta,\phi}\right|)
% .\label{point_to_line}
% \end{equation}
where $ {\Vec{\mathbf q_i}}$ represents the vector from $ {\Vec{\mathbf c}}$ to $ \Vec{\mathbf m}_i$, and each line $ \Vec{\mathbf l}_{\theta,\phi}$ is calculated by (\ref{line ejection}). Among those points, only the point farthest from $\Vec{\mathbf c}$ with Euclidean distance will be saved. Several points will inevitably appear multiple times in the final result. An iterative method is required to search and filter out. Afterward, we obtain a point cloud $E$, which only includes exterior points in the sampled point cloud $M$.\par

\subsection{K-means Region Segmentation}
\begin{figure}
    \centerline{\includegraphics[width=\columnwidth]{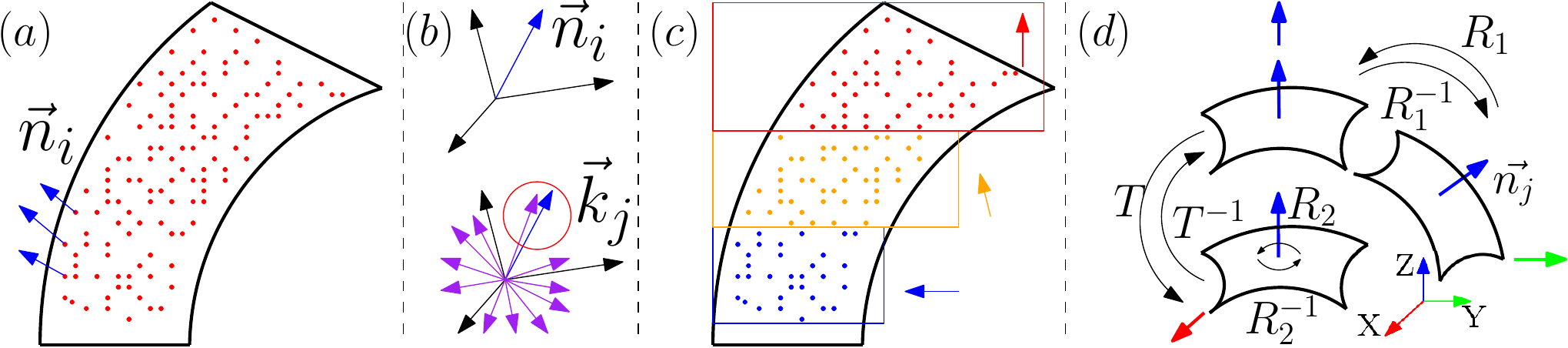}}
    \caption{Region segmentation using Algorithm \ref{kmeans_alg}. (a) Compute the normal of each point in the point cloud. (b) Initialize $K$ region normals randomly and find the most similar region normal for each point. (c) Re-compute the average normal of each region after one loop. (d) Region $R_j$ transformation.}
    \label{region}
\end{figure}

Normal is a fundamental feature of point cloud processing, as shown in Fig. \ref{region}(a). \cite{rusu20113d} PCL offers a function to compute the normal of each point. The normal of a point is a vector $ \Vec{\mathbf n}_i={[n_x \ n_y\ n_z]}^T$.
\par
In the field of point cloud segmentation, \cite{grilli2017review} provided a schematic representation, including edge-based, region growing, model fitting, hybrid method, and machine learning application. \cite{vo2015octree} presented a fast surface patch segmentation of urban environment 3D point clouds. This algorithm considers the geometry feature between adjacent points according to the concept of region growing. The normal variance of the region increases with its area.\par
We propose a region segmentation algorithm based on K-means clusters to meet the requirement of line scan camera, as shown in Algorithm \ref{kmeans_alg}. K-means cluster is a famous unsupervised classification algorithm in machine learning, which was originally presented by \cite{macqueen1967some}. When we consider a flat plane, the most outstanding feature is the same normal direction. However, finding a completely flat plane in a free-form object is very difficult. The solution here is trying to find the flat region whose largest normal angle difference is lower than a certain threshold. Fig. \ref{region}(b) and (c) illustrate the idea of the algorithm. For region $R_j$, we use the average normal to represent its normal:
\begin{equation}
  \Vec{\mathbf r}_j=\frac{\frac{1}{N_j}\sum_{i=1}^{N_j}{ {\Vec{ \mathbf n_i}}}}{||\frac{1}{N_j}{\sum_{i=1}^{N_j}{ \vec{\mathbf n}_i}}||}
.\label{average_normal}
\end{equation}
where $ {\vec{\mathbf n}_i}$ is the normal of i-th point in j-th region $R_j$, and $N_j$ is the size of region $R_j$. Now, $ {\vec{\mathbf r}_j}$ represents the average normal of region $R_j$. We can compute the angle between the normal of i-th point $\vec{\mathbf n}_i$ and the normal $\Vec{\mathbf r}_j$ of region $R_j$.  
\begin{equation}
\alpha_{ij}=\arccos{\frac{ {\vec{\mathbf n}_i} \cdot  {\vec{\mathbf r_j}}}{| {\vec{\mathbf n}_i}|\cdot| { \vec{\mathbf r}_j}|}}
.\label{vector_angle}
\end{equation}
For normal $\vec{\mathbf n}_i$, $K$ number of angles will be calculated. The smallest one will be emphasized and $\vec{\mathbf n}_i$ will be classified into the corresponding region $R_j$. After one loop. the estimated normal $ {\vec{\mathbf r}_j}$ for region $R_j$ will be computed again by (\ref{average_normal}). The points with similar normals will be classified as the same region according to the iteration loop.\par
In the normal K-means cluster algorithm, the only hyperparameter is the number of clusters, which is generally represented by $K$. Some useful methods are provided to guide us find the optimal $K$. However, setting the value of $K$ in advance with different free-form shapes of the object is impossible in our case. Instead, we will add $K$ iteratively until the condition of convergence is met. The convergence condition is defined as $\max \alpha_{ij}<\gamma$, where $\gamma$ is an angle threshold, which can be seen as the accuracy of the segmentation algorithm. For each loop, $\sigma_t$ is the number of regions that can meet the convergence condition. For each $K$, if the convergence of current loop $\sigma_t$ is smaller than the previous loop $\sigma_{t-1}$, which $t$ is the discrete time in each loop, the loop of $K$ is stopped. Then, the value of $K$ increases and the algorithm goes into another new loop. 
\begin{algorithm}
	\caption{K-means Region Segmentation}
	\label{kmeans_alg}
	\begin{algorithmic}[1]
	\Repeat
	    \State Initialize $K$ random normalized normal  vectors $\vec{\mathbf r}_j$
	    \While{$\sigma_t>\sigma_{t-1}$}
            \State Compute angle $\alpha_{ij}\leftarrow{(\ref{vector_angle})}$
            \State $j\xleftarrow{}{\arg\min}_{j}\alpha_{ij}$, push $E_{i}\;\rightarrow{R_j}$
            % \State push $E_{i}\;\rightarrow{R_j}$
            \State Update all $ {{\vec{\mathbf n}_j}}$  $\leftarrow{(\ref{average_normal})}$, Compute $\sigma_t$
            
            % \State Compute convergence number $\sigma_t$
         \EndWhile
         \State $K\leftarrow{K+1}$
    \Until {$\arg\max\alpha_{ij}<\gamma$}  
	\end{algorithmic} 
\end{algorithm}
\subsection{Adaptive ROI Path Planning}
\begin{figure}
    \centerline{\includegraphics[width=\columnwidth]{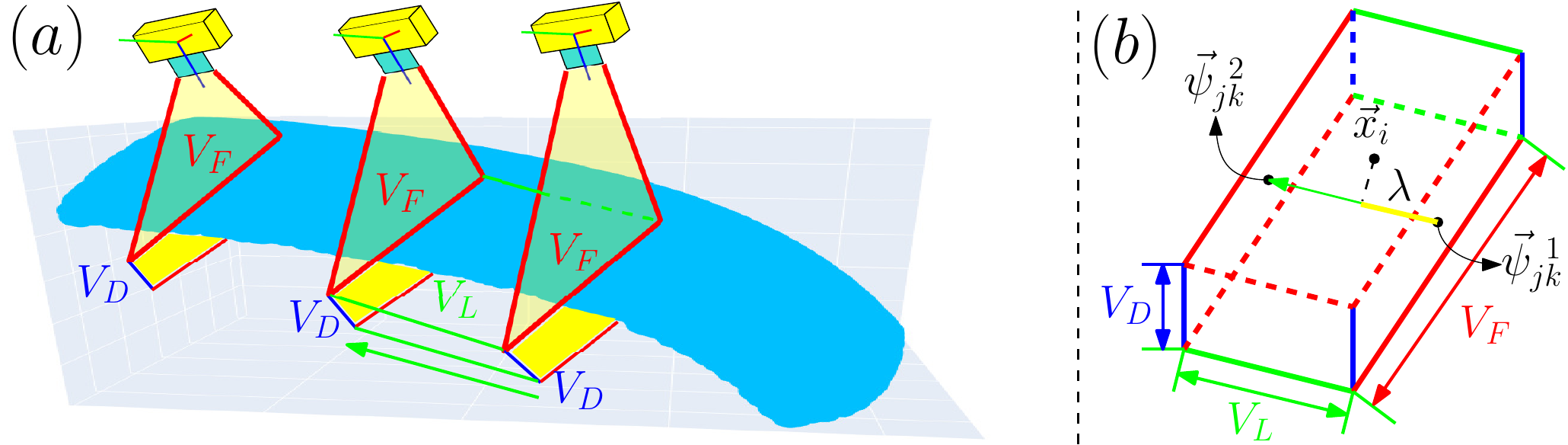}}
    \caption{ Completeness validation. $V_F,V_D,V_L$ represent the field of view, depth of view and moving of view respectively.}
    \label{completeness}
\end{figure}

\begin{figure}
    \centerline{\includegraphics[width=\columnwidth]{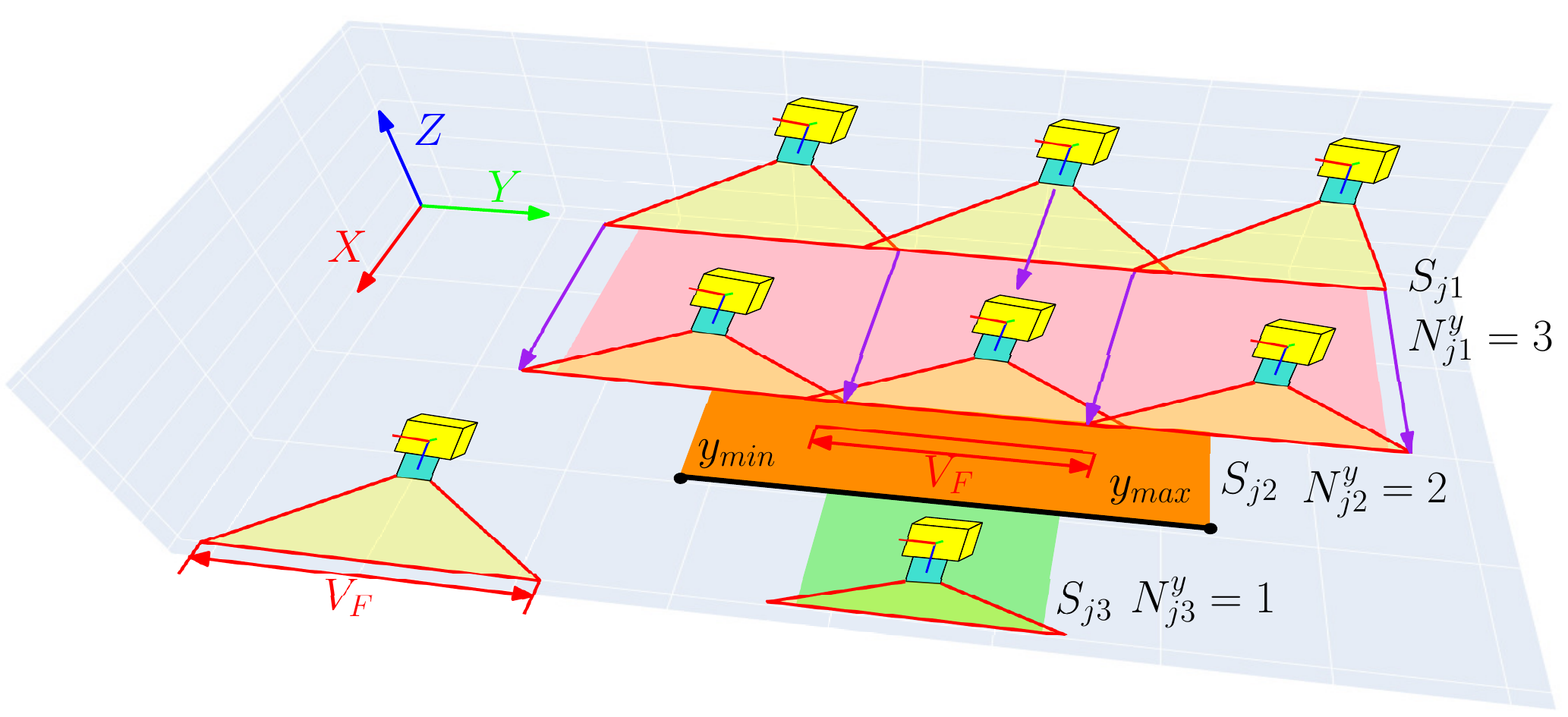}}
    \caption{$S_{j1}$, $S_{j2}$ and $S_{j3}$ represent three subregions in region $R_j$. $\Omega^y_{js}$ is the number of patches in subregion $S_{jk}$. }
    \label{adaptive_ROI_y}
\end{figure}

\begin{figure}
\centerline{\includegraphics[width=\columnwidth]{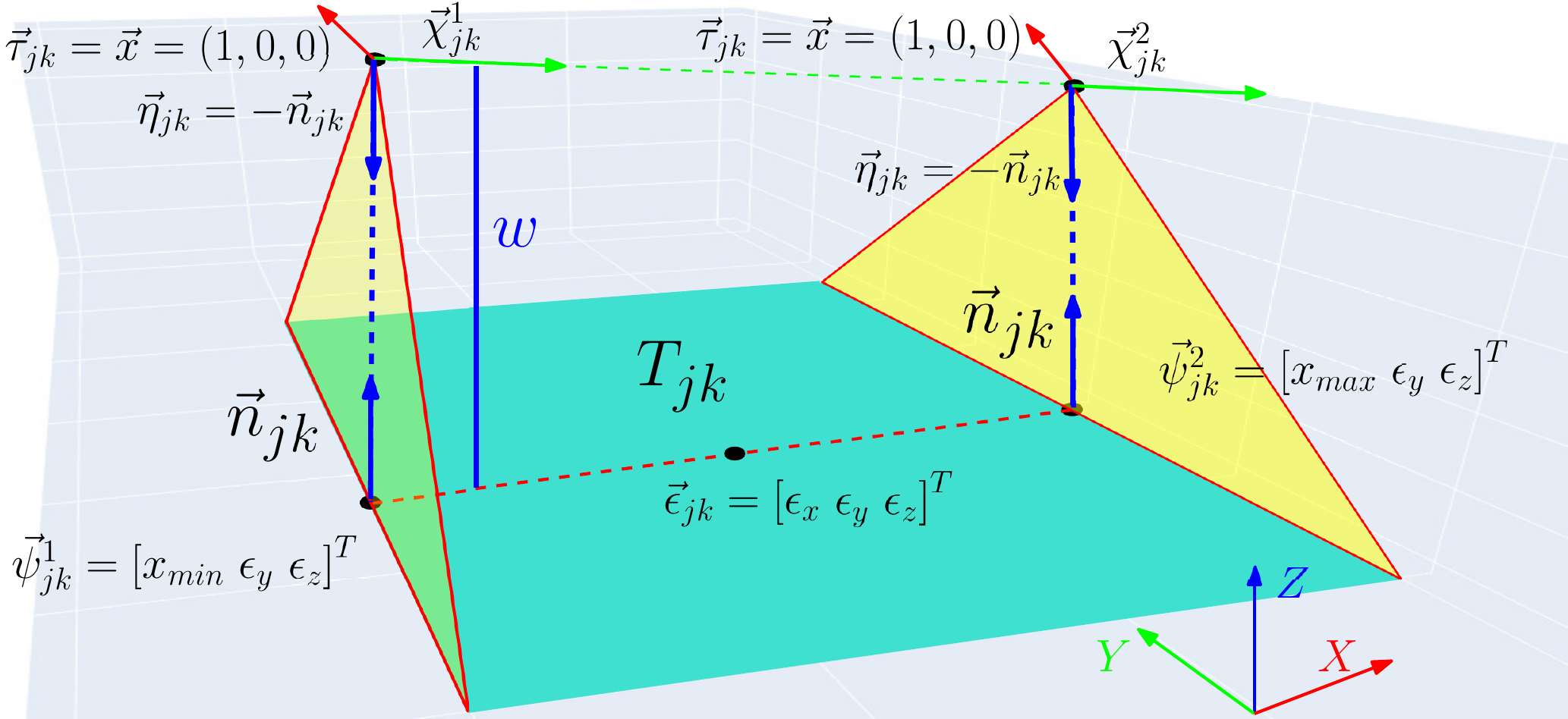}}
    \caption{A point cloud of patch $T_{jk}$. Two path point poses are illustrated by two frames in the figure. }
    \label{adaptive_ROI_position}
\end{figure}

After region segmentation, the shape of each region $R_j$ is various and irregular. This adaptive path planning algorithm takes the point cloud of each region $R_j$ as an input and output scanning path according to the characteristics of the line scan camera and common path planning requirements. We need to transform region $R_j$ to a suitable pose by three $4\times4$ transformation matrices at first, as shown in Fig. \ref{region}(d). The detailed explanation is:
\begin{itemize}
    \item [1)] 
    $\mathbf R_z:$ Rotate $ \vec{\mathbf n}_j$ to $ \vec{\mathbf z}=(0,0,1)$.
    \item [2)] 
    $\mathbf T_g:$ Translate to x-y plane.
    \item [3)] 
    $\mathbf R_p:$ Project to the plane
    $Z=0$ and extract the shortest straight line $ \vec{\mathbf b}_j$ in the boundary. Rotate $ \vec{\mathbf b}_j$ to $ \vec{\mathbf x}=(1,0,0)$.
\end{itemize}
Depth of view $V_D$, Field of view $V_F$ and normal similarity are three requirements for the line scan camera to scan the surface. When we consider the appearance inspection of the workpiece, completeness is a very fundamental requirement, because defects can appear anywhere. The idea of our adaptive ROI path is dividing region $R_j$ into several patches $T_{jk}$, in which each patch $T_{jk}$ is considered as a flat plane. As a result, the orientation of the camera remains the same during scanning patch $T_{jk}$. Fig. \ref{completeness}(a) illustrates the scanning model of a line scan camera. The rectangle formed by $V_F$ and $V_D$ is the coverage area of a scan line at a discrete time. When the line scan camera is moving in patch $T_{jk}$, the rectangle will be integrated as a standard cube. If all points in patch $T_{jk}$ are in the cube, the completeness is proven. The mathematical model of this cube is:
\begin{equation}
(   \vec{\mathbf t}_i-(  \vec{\boldsymbol \psi}_{jk}^{1}+\lambda \cdot \vec{\mathbf d})) \cdot  \vec{\mathbf d}=0 
\label{completness_equa}
\end{equation}
where ${\vec{\mathbf t}_i}$ is i-th point in patch $T_{jk}$, $\vec{\boldsymbol \psi}_{jk}^{1}={[x_{min}\ \epsilon_y\ \epsilon_z]}^T$ is illustrated in Fig. \ref{adaptive_ROI_position}, $\lambda$ is the projection length to the reference line, and $ \vec{\mathbf d}$ is the direction of the reference line. $\lambda$ should be smaller than $V_L$, $\frac{V_F}{2}$ and $\frac{V_D}{2}$ in three directions of the cube, as shown in Fig. \ref{completeness}(b).
Another requirement is normal similarity
 \begin{equation}
\max\alpha^i_{jk}<\beta
\label{judge}
\end{equation}
$\alpha^i_{jk}$ is the angle between i-th normal in patch $T_{jk}$ and average normal $\vec{\mathbf n}_{jk}$ of patch $T_{jk}$. $\beta$ is the allowable maximum angle threshold, decided by the machine vision system performance. Here, similar to K-means region segmentation, we divide region $R_j$ into $k_j$ subregions along the x-axis and $k_j$ is iteratively increased until all the conditions are met, as shown in Fig. \ref{adaptive_ROI_y}. Subregion $S_{js}$ is segmented into $\Omega^y_{js}$ number of patches.
\begin{equation}
\Omega^y_{js}=\lceil {(y_{max}-y_{min})}/{V_F} \rceil
\label{needed segments in y direction}
\end{equation}
where $\lceil x \rceil$ is ceiling function; $y_{max}$ and $y_{min}$ are the maximum and minimum values along the y-axis, respectively. The total sum of patches in region $R_j$ can be computed: 
\begin{equation}
\Omega_j=\sum_{s=1}^{k_j}{\Omega^y_{js}}
\label{needed segments}
\end{equation}

After patch $T_{jk}$ is divided from $R_j$, the position of two path points for it can be calculated, as shown in Fig. \ref{adaptive_ROI_position}: 
\begin{equation}
     \vec{\boldsymbol \chi}_{jk}^{1} =\vec{\boldsymbol \psi}_{jk}^{1} + w \cdot  \vec{\mathbf r}_{jk},\qquad
     \vec{\boldsymbol \chi}_{jk}^{2} =\vec{\boldsymbol \psi}_{jk}^{2} + w \cdot \vec{\mathbf r}_{jk}
\label{camera_position}
\end{equation}
where $\vec{\boldsymbol \psi}_{jk}^1={[x_{min}\ \epsilon_y\ \epsilon_z]}^T$ and $\vec{\boldsymbol \psi}_{jk}^2={[x_{max}\ \epsilon_y\ \epsilon_z]}^T$,  $x_{max}$ and $x_{min}$ are the maximum and minimum values of patch $T_{jk}$ along the x-axis, respectively. $\epsilon_y$ and $\epsilon_z$ are elements of $ \vec{\boldsymbol \epsilon}_{jk}$, computed by (\ref{center_point}) respect to $T_{jk}$. $w$ denotes the working distance, the distance between the camera and the surface; and $ \vec{\mathbf r}_{jk}$ is the average normal of patch $T_{jk}$ computed by (\ref{average_normal}). Here, the normal direction of camera is $\vec{\boldsymbol \eta}_{jk}=-\vec{\mathbf r}_{jk}$. In addition, the moving direction of the line scan camera for patch $T_{jk}$ is $  \vec{\boldsymbol \tau}_{jk}= \vec{\mathbf x}=(1,0,0)$. The orientation of the camera is unique and solvable with these two defined directions. Now, $P_{jk}$ is the path for scanning a patch $T_{jk}$ and the i-th six-DOF path point pose $\vec{\mathbf p}^i_{jk}$ of $P_{jk}$ can be defined as:
\begin{equation}
 \vec{\mathbf p}^i_{jk}=\left\{ \mathbf T_{inv} \cdot \vec{\boldsymbol \chi }^i_{jk}, \mathbf R_{inv} \cdot \vec{\boldsymbol \eta }^i_{jk}, \mathbf R_{inv} \cdot \vec{\boldsymbol \tau} ^i_{jk}\right\}
\label{path pose}
\end{equation}
where $\mathbf T_{inv}=\mathbf R_z^{-1} \cdot\mathbf {T_g}^{-1} \cdot \mathbf R_p^{-1}$ and $\mathbf R_{inv}=\mathbf R_z^{-1} \cdot \mathbf R_p^{-1}$. $\vec{\boldsymbol \chi }^i_{jk}$, $\vec{\boldsymbol \eta }^i_{jk}$ and $\vec{\boldsymbol \tau} ^i_{jk}$ are $4\times1$ vectors in homogeneous coordinates.
\par
After computing the scanning path $P_{jk}$ for patch $T_{jk}$, we can stack them as a complete scanning path $P_j$ for region $R_j$. We need to integrate the path $P_j$ of different regions into a whole scanning path to obtain a complete scanning path for the specular object. Here, we adopt an intuitive distance optimal path. The selected principle is like the nearest neighbor search. The distance used is Euclidean distance among different $\vec{\boldsymbol \chi }^i_{jk}$ . The principle is always searching the nearest region in the rest of unscanned regions as the next one after scanning one region. Following this principle, we can achieve a complete path for scanning the whole free-form object.

\section{SENSING AND CONTROL SYSTEM}
\subsection{Projection Registration for Specular Surfaces}
In the above chapter, the scanning path is computed in the object's frame. We need a transformation matrix for the robot frame to control the manipulator. This transformation matrix is denoted as:
\begin{equation}
 ^{B}\mathbf T_{O}=\left[\begin{array}{cc}
\mathbf R_B |   {\vec{\mathbf t}_B}\\
\end{array}
\right] 
\label{T_BO}
\end{equation}
where $ ^{B}\mathbf T_{O}$ maps object frame to robot frame. $\mathbf R_B$ is a $3\times3 $ rotation matrix and $ {\vec{\mathbf t}_B}$ is a $3\times1$ translation column vector.\par
\begin{figure}
    \centerline{\includegraphics[width=\columnwidth]{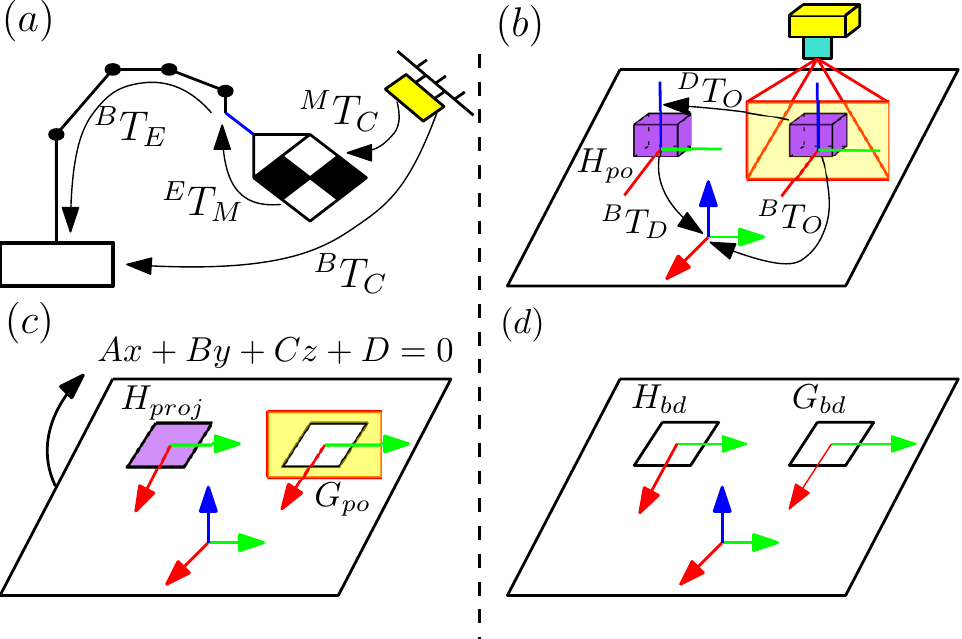}}
    \caption{The pipeline of the projection registration algorithm: (a) Hand-eye calibration. (b) Real experiment set up. (c) Projection point cloud $H_{proj}$ and capturing point cloud $G_{po}$. (d) Two boundary point clouds $H_{bd}$ and $G_{bd}$.}
    \label{projection_registration}
\end{figure}

The recent development of the RGB-D camera made it an effective device to obtain 3D information in the space. Thus, a commercial RGB-D camera is also included in our proposed system. A hand eye calibration should be conducted in advance to describe the relationship between the robot and the camera, as shown in Fig. \ref{projection_registration}(a):
\begin{equation}
 ^B \mathbf T_C= ^B \mathbf T_E\cdot  ^E \mathbf T_M \cdot  ^M \mathbf T_C
\label{handeye calibration}
\end{equation}
where C represents the RGB-D camera, B is the robot base, E represents the end-effector of the robot manipulator, and M represents AR markers on the calibration board. $^B \mathbf T_E$ can be achieved by robot forward kinematics. $^E \mathbf T_M$ is determined by our designed calibration board. $^M \mathbf T_C$ is detected by the RGB-D camera via \cite{ar_track_alvar}. \par
Nowadays, there are different types of RGB-D camera, such as TOF, structured light and stereo vision. However, none of them can capture the information of specular surface due to the loss of reflection light. Thus, we propose a projection registration method to deal with this issue. Fig. \ref{projection_registration}(b) illustrates the setting of our system, where an RGB-D camera is mounted on top of the workspace. The left one $H_{po}$ denotes the designed pose of the object in the robot's frame. The right one is the actual pose of the object in the robot's frame. The RGB-D camera can capture the point cloud of manipulation plane $G_{po}$, where a hole appears because it is blocked by the specular surface of the object, as shown in Fig \ref{projection_registration}(c). The left one $H_{proj}$ is the projection of $H_{po}$. The manipulation plane in the robot frame can be defined as $A\cdot x+B\cdot y+C\cdot z+D=0$. Then, the point cloud projected to this parametric plane is:
\begin{equation}
t=  {\vec{\mathbf f}} \cdot   {\vec{\mathbf h}_i^{po}},\qquad
  {\vec{\mathbf h}_i^{proj}}=  {\vec{\mathbf h}_i^{po}}-t\cdot   {\vec{\mathbf f}}
\label{projection coordinate}
\end{equation}
where $\vec{\mathbf f}=[A\ B\ C\ D]^T $ is the normal vector of the parameteric plane; t is a reference variable; $ \vec{\mathbf h}_i^{po}$ and  $\vec{\mathbf h}_i^{proj}$ are i-th point in homogenous coordinates of $H_{po}$ and $H_{proj}$ respectively. Given $H_{proj}$ and $G_{po}$, we can extract their boundaries to compute their transformation matrix. \cite{rusu20113d} offered a function to extract the boundary of $H_{proj}$, denoted as $H_{bd}$. For $G_{po}$, we adopt a method similar to Fig. \ref{preprocessing}(c) to achieve that, denoted as $G_{bd}$. Finally, we can use the classical ICP \cite{rusinkiewicz2001efficient} algorithm to derive the least square transformation matrix $^D \mathbf T_O$ between $H_{bd}$ and $G_{bd}$. Then, $^{B}\mathbf T_{O}$ can be obtained by $^{B}\mathbf T_{O}=^{B}\mathbf T_{D}\cdot^{D}\mathbf T_{O}$. 

\subsection{Automatic Scanning}
The scanning path in the robot frame can be obtained by
\begin{equation}
 \vec{\mathbf p}_B=\left\{ ^B\mathbf T_O \cdot { \vec{\boldsymbol \chi }}_O,\mathbf R_B\cdot {\vec{\boldsymbol \eta}} _O,\mathbf R_B\cdot {\vec{\boldsymbol \tau}}_O \right\}
\label{path pose in robot frame}
\end{equation}
where $\vec{\boldsymbol \eta}^O$ and $ \vec{\boldsymbol \tau}^O$ are $3\times1$ column vectors, and $ \vec{\boldsymbol \chi }_O$ is a $4\times1$ column vector in homogeneous coordinates. Every pose should be a $6\times1$ column vector, representing the position and the rotation vector in the robot frame to control the robot. $  {\vec{\boldsymbol \chi}_B}$ can directly define the desired position of the robot end-effector. The entire transformation matrix must be calculated based on our defined direction of $  \vec{{\boldsymbol \eta}_B}$ and $  \vec{\boldsymbol \tau}_B$ to define the orientation of the robot end-effector. Then, we can obtain the rotation vector with \cite{baker2012matrix}, denoted as $  \vec{\boldsymbol \xi}_B$. The pose vector is as follows to define the pose of the robot end-effector for each path point:
\begin{equation}
  {\vec{\boldsymbol \pi}}=\left[\begin{array}{cc}
{  {\vec{\boldsymbol \chi}}^T} & {  \vec{{\boldsymbol \xi}}^T} \\
\end{array}
\right] ^T
\label{path pose vector}
\end{equation}
where ${\vec{\boldsymbol \pi}}$ is a $6 \times 1$ column vector.\par
We can control the camera to finish the automatic inspection task with the scanning path in the robot frame. Several substantial optics parameters need to be calculated before conducting the machine vision inspection task. The lens used in our platform is Cinegon 1.8/16-0901. $V_F$ and $V_D$ of the camera can be calculated according to \cite{burke2012image}:
\begin{equation}
V_F =2 w\tan({\Theta/2}),\quad c \approx \delta/1500,\quad V_D \approx 2w^2 f'c/f^2
\label{FOV}
\end{equation}
where w denotes the working distance, $\Theta$ denotes the angel of view, $\delta$ denotes the sensor diagonal, c denotes the circle of confusion, $f'$ denotes the f-number, and $f$ denotes the focal length.\par
\begin{figure}
    \centerline{\includegraphics[width=\columnwidth]{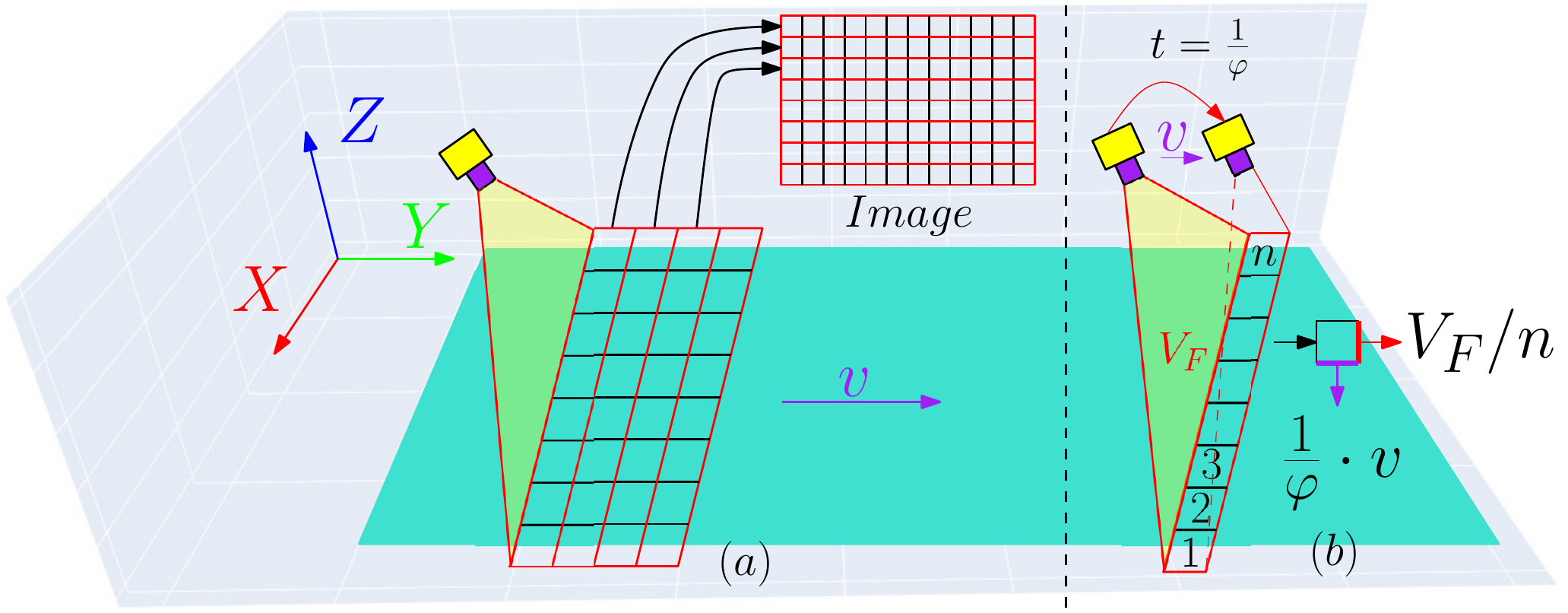}}
    \caption{Automatic scanning. (a) Images captured during line scan motion. (b) Acquisition line rate with unit magnification.}
    \label{automatic_scanning}
\end{figure}

The scanning of the line scan camera is illustrated in Fig. \ref{automatic_scanning}(a). Pixels in each scanning line are saved in the buffer continuously with the motion of the line scan camera. Acquisition line rate is one of the most important parameters of the line scan camera, which is needed to match with the moving speed of the camera, with this principal. Otherwise, the shapes in the captured images will be distorted. Magnification parameter must be equal to one, which describes the ratio between the pixel size and the actual object size. The geometric relationship of the pixels to achieve a unit magnification is shown in Fig. \ref{automatic_scanning}(b). Thus, we can obtain
\begin{equation}
V_F/n=(1/\varphi)\cdot v
\label{line rate}
\end{equation}
where $n$ is the number of pixels of the image, $\varphi$ is the acquisition line rate, and $v$ is the moving speed of the robot.\par
\subsection{Image Mapping}
\cite{opencv_library} OpenCV is used for image processing over the captured images. Gaussian smoothing is used to filter out the noise of the images. Next, Canny edge operator \cite{canny1986computational} is used to detect the defects in the image.\par
The detection results in 2D image information can be mapped to the 3D model according to the corresponding manipulator scanning pose. Here, we also use the point cloud to represent the texture on the specular surface of the object. In the above adaptive ROI path planning, each patch $T_{jk}$ is regarded as a flat plane. As a result, the distance between the surface and the line scan camera remains the same during scanning patch $T_{jk}$. The first path pose $\vec{\boldsymbol \pi}^1_{jk}$ of patch $T_{jk}$ is used as our mapping frame. Fig. \ref{mapping} illustrates the geometric mapping relationship. The mathematical model is:
\begin{figure}
    \centerline{\includegraphics[width=\columnwidth]{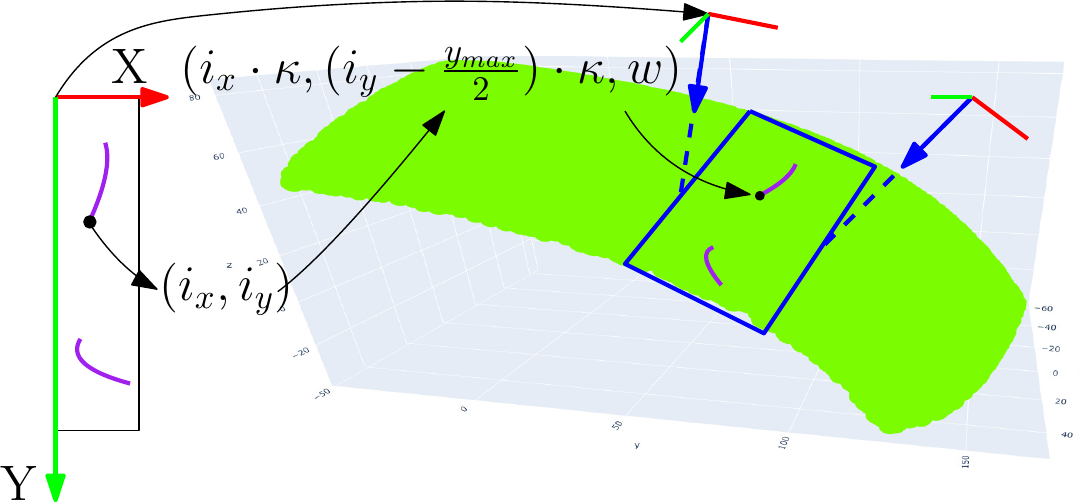}}
    \caption{ Image mapping. The pixel point in image frame (in the left) can be mapped to object 3D frame (in the right).}
    \label{mapping}
\end{figure}
\begin{equation}
  {\vec{\boldsymbol \mu}}={\left[\begin{array}{ccc}
i_x\cdot \kappa & (i_y-y_{max}/2)\cdot \kappa & w \\
\end{array}
\right]}^T 
\label{image mapping}
\end{equation}
where $i_x$ and $i_y$ are elements of $\vec{{\mathbf i}}=(i_x,i_y)$ in the 2D image frame, $y_{max}$ is the maximum value along the y-axis in the 2D image frame, $\kappa$ represents the pixel size in the image, and ${\vec{\boldsymbol \mu}}$ is a $3\times1$ vector representing the position of the defects in the local frame $\vec{\boldsymbol \pi}^1_{jk}$. 
% These points can be transformed to the global object frame according to the scanning pose $\vec{\boldsymbol \pi}^1_{jk}$.

\section{RESULTS}
\subsection{Experiment Setup}
\begin{figure}
    \centerline{\includegraphics[width=0.9\columnwidth]{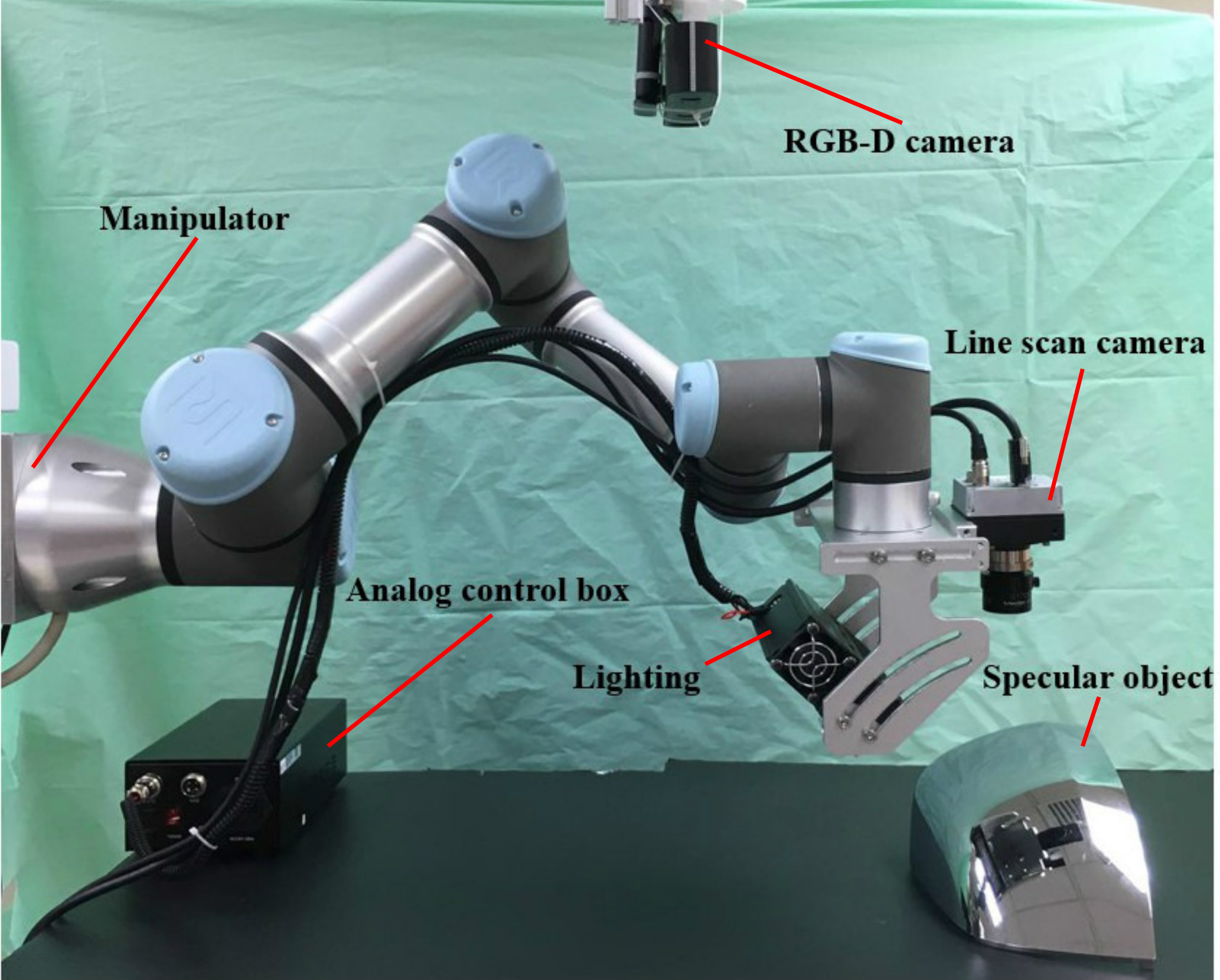}}
    \caption{Experiment set up of the inspection task.}
    \label{experiment_setup}
\end{figure}

The experimental setup of our system is illustrated in Fig. \ref{experiment_setup}. An Orbbec astra RGB-D camera is mounted on top of the manipulation space and used to capture the 3D point cloud of the manipulation plane. The transformation matrix between the RGB-D camera and the UR3 manipulator is calibrated by ArUco markers in advance. Velocity and acceleration of the manipulator are empirically set as $0.05m/s$ and $0.05m/s^{-2}  $ due to safety reasons. Working distance $w$ of the line scan camera is set as $15.7mm$ according to the focal length of the len. As a result, acquisition line rate can be computed as $670\ line/s$. The machine vision system configuration serves as dark-field illumination, in which the angle between the high-intensity line lighting and the line scan camera needs to be adjusted through several tests. This angle should guarantee that only the high intensity light reflected on the uneven scratches can reach the camera view. A specular side mirror from the automotive industry is taken as an experimental workpiece, placed on the manipulation plane. Here, several defects are added manually on the specular surface and the observation results from professional workers are regarded as the benchmark of this inspection task.\par
Three vital elements are considered to evaluate the performance of the proposed system:
\begin{enumerate}[]
\item Robustness and computation efficiency of path planning with respect to different free-form shapes.
\item Error between the projection boundary of the CAD model and the detected boundary after alignment.
\item Accuracy of defects detection accuracy over the captured images.
\end{enumerate}

\subsection{Path Planning Performance}
\begin{figure}
    \centerline{\includegraphics[width=\columnwidth]{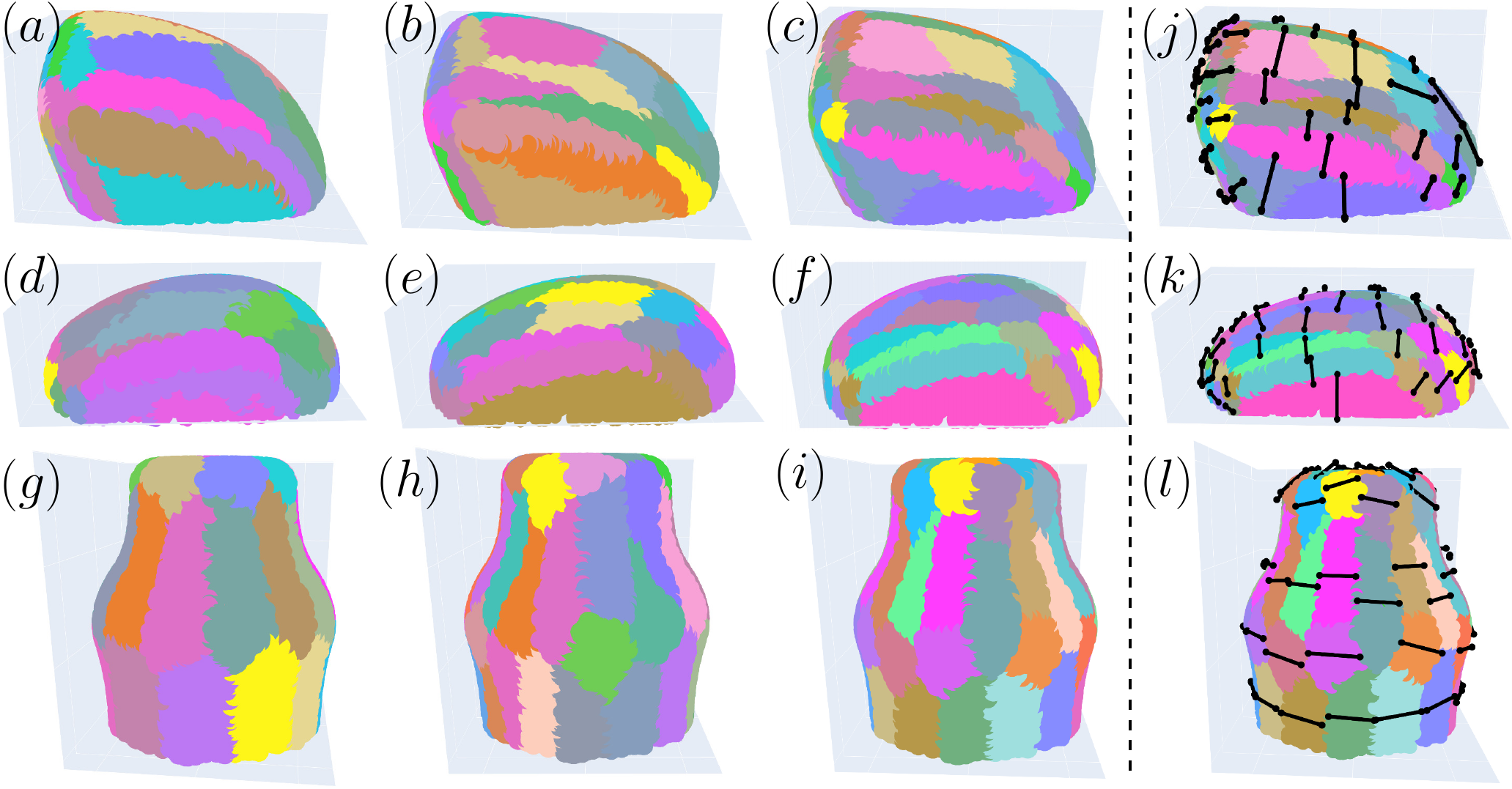}}
    \caption{(a)-(c), (d)-(f) and (g)-(i) are region segmentation results of side mirror, mouse and bottle respectively with different angle thresholds $\gamma$. (j)-(l) are simplified adaptive paths for regions respect to side mirror, mouse and bottle respectively.
    }
    \label{results_path_performance}
\end{figure}

The proposed path planning system is implemented on a Linux-based PC and, written in C++. One of the largest challenges of our task is the variety of object shapes. The CAD mesh models of a side mirror, a mouse, and a bottle are applied in this section to validate the robustness in terms of different free-form shapes. In addition, different input parameters are tested to compare their results. Fig. \ref{results_path_performance} (a)-(i) shows the different region segmentation results with respect to different free-form objects and input parameters. We can conclude from these intuitive results that our K-means unsupervised region segmentation can be suitable for different convex free-form objects.
The mathematical results, including the number of regions and efficiency, are shown in Table \ref{path planing performance}. After region segmentation, adaptive ROI path planning can be conducted. Adaptive ROI means that regions with arbitrary shapes can be divided into suitable patches to be completely scanned by the line scan camera. The path of different patches belonging to the same region is integrated in Fig. \ref{results_path_performance} (j)-(l) to visualize the path clearly. This path planning can compute optimal scanning directions for different shape areas.

\begin{table}
% [!htbp]
\centering
\caption{Path Planning Performance}
\label{path planing performance}
\begin{tabular}{|c|c|c|c|c|}
%\begin{tabularx}{\textwidth}{|c|c|c|c|c|}
\hline
Object & $N_E$ & $\gamma$ & $N_R$ & T(s)  \\
\hline
\multirow{3}*{Side Mirror} & \multirow{3}*{4807} & 20 & 30 & 12 \\
\cline{3-5}
~ & ~ & 18 & 36 & 17  \\
\cline{3-5}
~ & ~ & 15 & 45 & 26  \\
\hline

\multirow{3}*{Mouse} & \multirow{3}*{3855} & 20 & 39 & 11 \\
\cline{3-5}
~ & ~ & 18 & 40 & 17  \\
\cline{3-5}
~ & ~ & 15 & 58 & 34  \\
\hline

\multirow{3}*{Bottle} & \multirow{3}*{7823} & 20 & 42 & 23 \\
\cline{3-5}
~ & ~ & 18 & 55 & 34  \\
\cline{3-5}
~ & ~ & 15 & 75 & 68  \\
\hline

\hline
\end{tabular}
\begin{tablenotes}
\item $N_E:$ size of exterior point cloud. $\gamma:$ angle threshold in region segmentation. $N_R:$ number of regions. $T(s):$ computation time for K-means region segmentation. 
\end{tablenotes}
\end{table}

\subsection{Registration}
\begin{figure}
    \centerline{\includegraphics[width=\columnwidth]{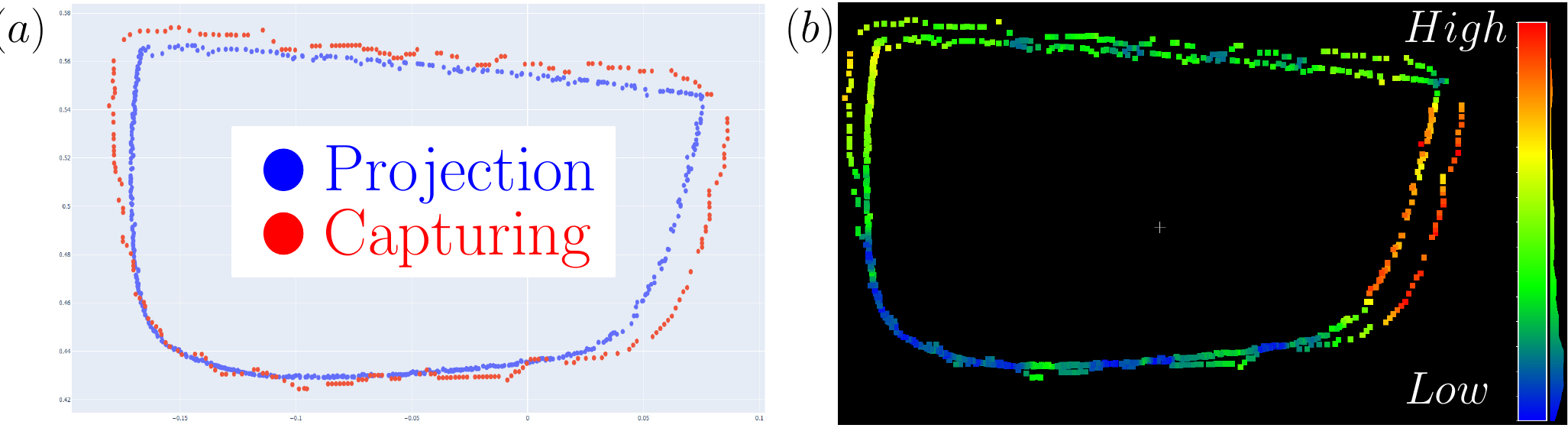}}
    \caption{Results of projection registration for specular surfaces. (a) Intuitive comparison between CAD mesh projection boundary and RGB-D camera capture boundary. (b) C2C absolute distances between them.}
    \label{results_registration}
\end{figure}

Registration means associating sets of data into a common coordinate. In our case, we need to obtain the transformation matrix $^B T_O$ to send motion command to the robot. A projection boundary registration method is proposed in our system to deal with the problem of losing the 3D point cloud on the specular surface. An Orbbec Astra 3D camera is used in our system, whose uncertainty in depth measurements is bounded by a $\pm 15 mm$ at a $2500 mm$ range \cite{giancola2018survey}. We compare the boundary of clean CAD model projection, which is considered ground truth, and the boundary captured by the RGB-D camera with our proposed processing algorithm after alignment by ICP to evaluate the accuracy of this method. The intuitive comparison results are shown in Fig. \ref{results_registration}(a). \cite{girardeau2015cloud} is used to compare two point clouds mathematically, as shown in Fig. \ref{results_registration}(b). The errors mean and standard deviation are $4.842\times10^{-3}$ and $3.548\times10^{-3}$ respectively. Such low errors can guarantee the computation scanning path can deal with the continuous curvature specular surface after the transformation.

\subsection{Detection Results}
\begin{figure}
    \centerline{\includegraphics[width=\columnwidth]{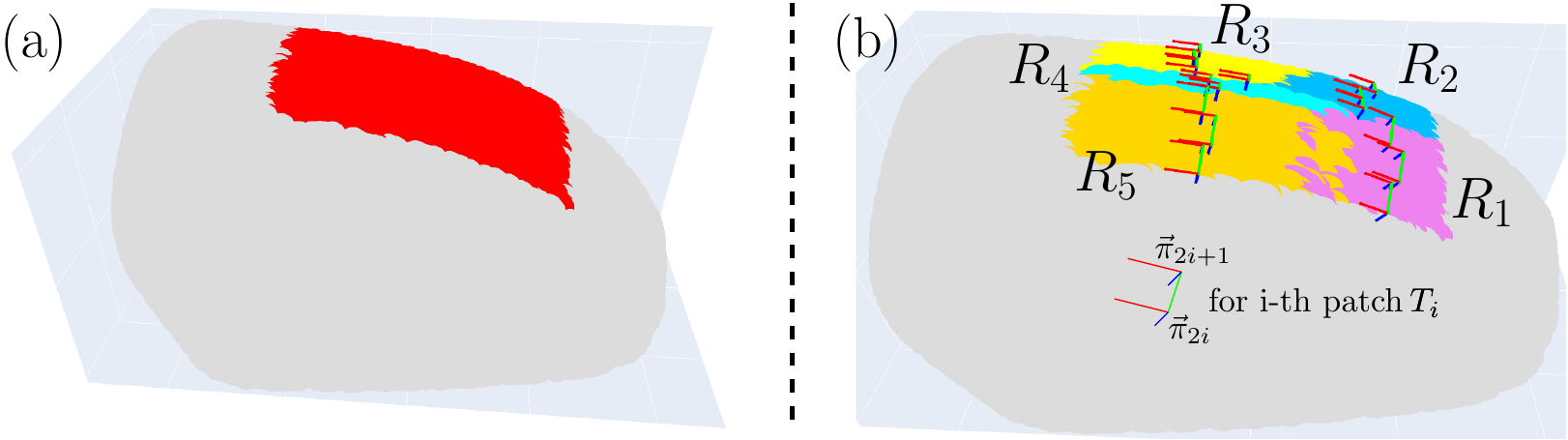}}
    \caption{Robotic experiment subject. (a) The chosen part of the specular side mirror. (b) The scanning path respect to the chosen surface. }
    \label{results_experiment}
\end{figure}

After the scanning path computation and projection registration, the robot and the line scan camera are controlled to conduct the inspection task. The image capture for patch $T_{jk}$ starts at $\vec{\boldsymbol \pi}_{2\cdot i}$ and stops at $\vec{\boldsymbol \pi}_{2\cdot i+1}$, where $i$ represents i-th patch of the surface because the line scan camera needs to cooperate with robot motion to capture image. Thus, the number of total images equals the number of patches. Constrained by the operational space of the UR3 manipulator, a small part of the specular side mirror is selected to conduct the experiments to illustrate the idea and validate our methods, as shown in Fig. \ref{results_experiment}(a). Fig. \ref{results_experiment}(b) illustrates the region segmentation result and scanning path.\par 
\begin{figure}
    \centerline{\includegraphics[width=\columnwidth]{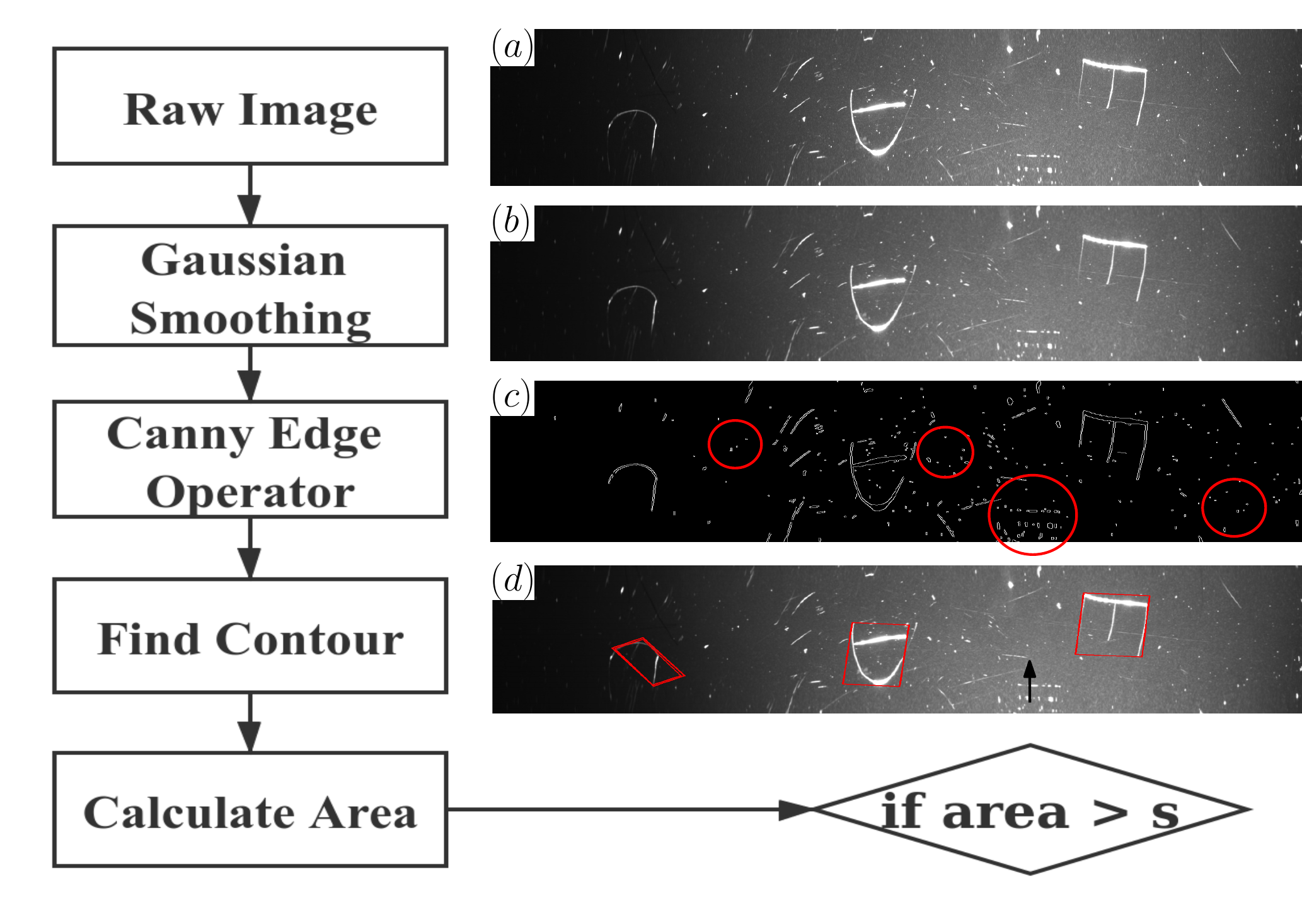}}
    \caption{Image processing pipeline and its corresponding results (a)-(d). }
    \label{image_processing_pipeline}
\end{figure}

\begin{figure}
     \centerline{\includegraphics[width=\columnwidth]{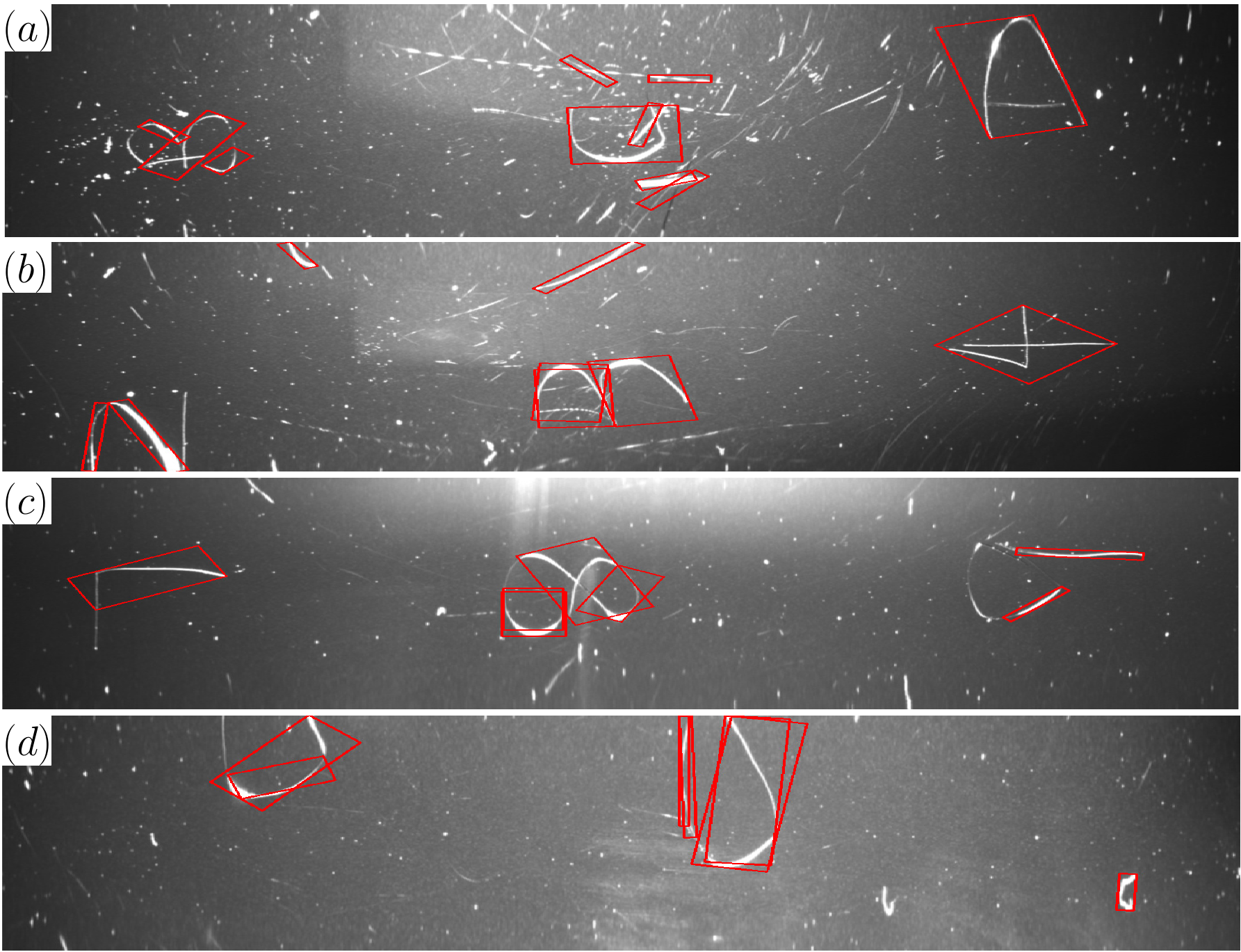}}
    \caption{(a)-(d) The captured images of the line scan camera and corresponding detection results. }
    \label{results_detection}
\end{figure}

Some dusts will be captured in the image and are considered edges in Canny operators due to the high resolution of the line scan camera, as shown in Fig. \ref{image_processing_pipeline} (a)-(c). However, the small dusts will not be considered unqualified defects in quality control. Thus, deeper image processing is needed to deal with this issue. A contour approximation method is adopted to find the contours in binary Canny processed images. Then, OpenCV also provides a function to calculate the area of the contours. A threshold $s$ can be tuned to judge if the contour is a defect or not, as shown in Fig. \ref{image_processing_pipeline} (d). Detection accuracy can be improved greatly through this algorithm. Fig. \ref{results_detection} (a)-(d) illustrate other results with this image processing algorithm. The red squares represent the detected defects in the images. Inspired by the classification problem in machine learning, precision, recall, and F-measure are used to evaluate the performance of this algorithm.
\begin{equation}
\begin{split}
    \rho=TP/(TP+FP),\quad& \upsilon=TP/(TP+FN) \\
    \iota =2\cdot (\rho \cdot \upsilon)&/(\rho+\upsilon)
\end{split}
\label{precision}
\end{equation}
where $\rho$ represents precision, $\upsilon$ denotes recall, and $\iota$ represents F-measure. The accuracy of the algorithm with respect to different regions is shown in Table \ref{detection_results}. The results reveal that our system performs well in defect detection for the specular surface. More tests in real production line are needed for further improvement. \par
\begin{figure}
    \centerline{\includegraphics[width=\columnwidth]{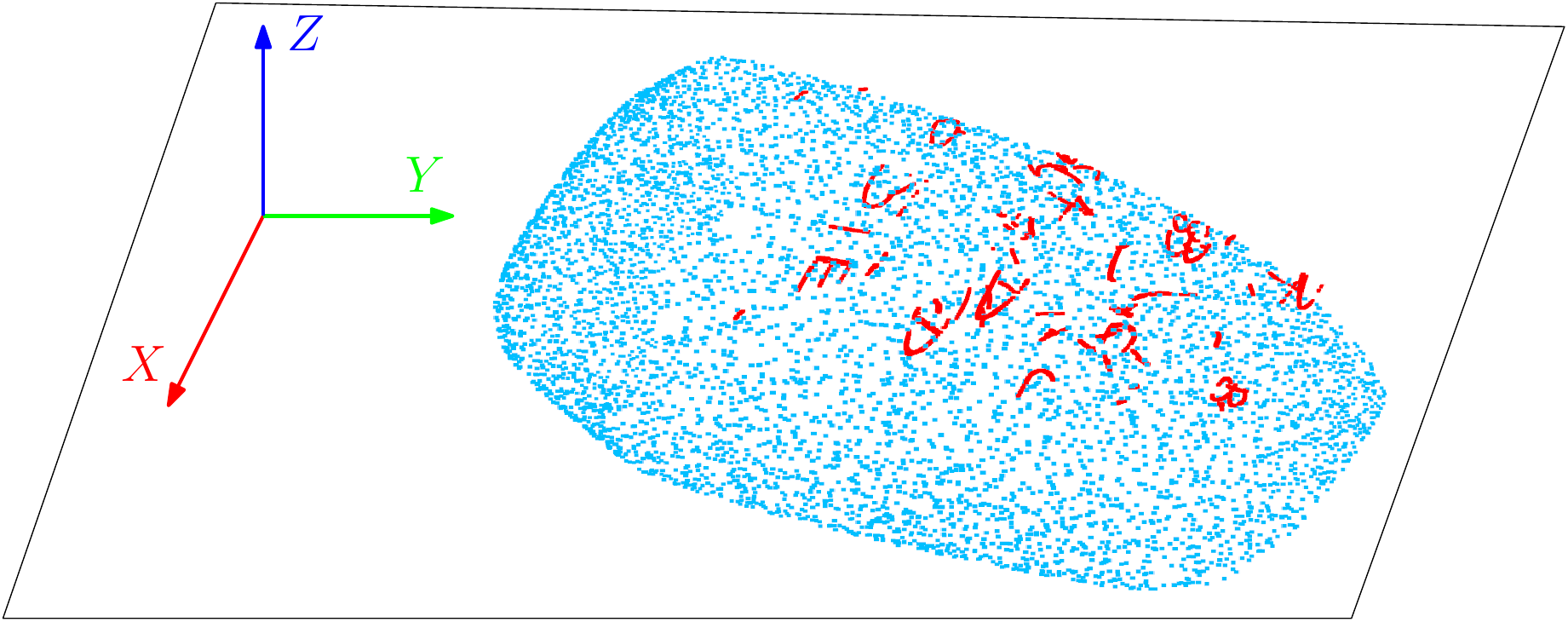}}
    \caption{3D mapping results according to defect detection. }
    \label{results_mapping}
\end{figure}

\begin{table}
% [!htbp]
\centering
\caption{Defect Detection Results}
\label{detection_results}
\begin{tabular}{|c|c|c|c|c|c|c|}
\hline
Part & TP & FP & FN & Precision & Recall & F-meas\\
\hline
$R_1$ & 20 & 3 & 0 & 0.870 & 1.000 & 0.930 \\
\hline
$R_2$ & 9 & 3 & 0 & 0.750 & 1.000 & 0.857 \\
\hline
$R_3$ & 11 & 1 & 2 & 0.917 & 0.846 & 0.880 \\
\hline
$R_4$ & 12 & 4 & 3 & 0.750 & 0.800 & 0.774 \\
\hline
$R_5$ & 10 & 5 &0 & 0.667 & 1.000 & 0.800 \\
\hline
\end{tabular}
\begin{tablenotes}
\item TP is true positive, FP is false positive, and FN is false negative. 
\end{tablenotes}
\end{table}
Based on the above discussion, each captured image belongs to patch $T_{jk}$. According to starting pose $\vec{\boldsymbol \pi}_{2\cdot i}$, the detected results in image $I_i$ can be mapped to i-th patch $T_{jk}$. The entire mapping result is shown in Fig. \ref{results_mapping}. We assume that patch $T_{jk}$ is totally flat, although it is not completely the case. This mapping can provide an intuitive distribution about the defects on the specular object.
% In this mapping, we assume that the patch $T_{jk}$ is totally flat. However, it is not real in the real surface of the object.  Thus, some errors are unavoidable in this mapping result.

\section{CONCLUSIONS}
In this paper, we demonstrate a robotic line scan system for inspection over convex free-form specular surfaces. The line scan camera can finish the defects inspection task over all sides of convex free-form objects with the help of matched lighting through the combination of machine vision and robotic manipulator. This system breaks the limitation that the previous common system can purely deal with standard-shaped objects. Simultaneously, the scanning performance in terms of specular curvature continuous surface is improved in our proposed system. \par
Taking a CAD mesh model as input, K-means based region segmentation and adaptive ROI path planning are used to compute the scanning path in terms of the sampled point cloud. A projection registration method is proposed to localize the object in a robot's frame. The sampling line rate of the line scan camera is matched with the manipulator's motion to finish the automatic inspection task. Experiments are carried out to validate the proposed system with an image processing pipeline. The experimental results show that our proposed system performs well in defect inspection.\par
% Currently, a manipulator with small working space is adopted in our system, limiting the scanning range of the line scan camera. 
Currently, we are looking for industry partners to test our proposed system in the production line. Furthermore, image processing algorithms, such as CNN in deep learning, need to be adopted to judge the defects and classify them into different kinds in the future.

\addtolength{\textheight}{-12cm}   % This command serves to balance the column lengths

%%%%%%%%%%%%%%%%%%%%%%%%%%%%%%%%%%%%%%%%%%%%%%%%%%%%%%%%%%%%%%%%%%%%%%%%%%%%%%%%

%%%%%%%%%%%%%%%%%%%%%%%%%%%%%%%%%%%%%%%%%%%%%%%%%%%%%%%%%%%%%%%%%%%%%%%%%%%%%%%%

%%%%%%%%%%%%%%%%%%%%%%%%%%%%%%%%%%%%%%%%%%%%%%%%%%%%%%%%%%%%%%%%%%%%%%%%%%%%%%%%

%%%%%%%%%%%%%%%%%%%%%%%%%%%%%%%%%%%%%%%%%%%%%%%%%%%%%%%%%%%%%%%%%%%%%%%%%%%%%%%%

% References are important to the reader; therefore, each citation must be complete and correct. If at all possible, references should be commonly available publications.

% \begin{thebibliography}{99}
% \end{thebibliography}
\bibliographystyle{ieeetr} %ieeetr国际电气电子工程师协会期刊
\bibliography{ref} % ref就是之前建立的ref.bib文件的前缀

\end{document}